\documentclass{article}

\usepackage[numbers,sort&compress]{natbib}

\usepackage[preprint]{neurips_2026}

\usepackage[utf8]{inputenc} 
\usepackage[T1]{fontenc}    
\usepackage{hyperref}       
\usepackage{url}            
\usepackage{booktabs}       
\usepackage{amsfonts}       
\usepackage{nicefrac}       
\usepackage{microtype}      
\usepackage{xcolor}         
\usepackage{graphicx}
\usepackage{amsmath}
\usepackage[table]{xcolor}
\usepackage{array}
\usepackage{wrapfig}
\usepackage{caption}
\usepackage[most]{tcolorbox}

\definecolor{promptnavy}{HTML}{2B4F7E}   
\definecolor{promptblue}{HTML}{4A7BB7}   
\definecolor{promptsky}{HTML}{EEF4FA}    

\definecolor{coral}{HTML}{E8735A}
\definecolor{ftnavy}{HTML}{184498}
\definecolor{fmpurple}{HTML}{C25DCB}
\definecolor{sfsgreen}{HTML}{3A7D3A}

\definecolor{citeblue}{HTML}{1055C9}

\hypersetup{
    colorlinks=true,
    linkcolor=citeblue,
    citecolor=citeblue 
}

\newtcolorbox{promptbox}[1]{
  enhanced,
  colback=promptsky,
  colframe=promptnavy,
  colbacktitle=promptnavy,
  coltitle=white,
  fonttitle=\bfseries,
  title=#1,
  arc=2mm,
  boxrule=0.8pt,
  left=3mm, right=3mm, top=2mm, bottom=2mm,
  breakable,
  before skip=8pt,
  after skip=8pt,
}

\newcommand{\role}[1]{\textbf{\color{promptnavy}[#1]}}

\usepackage[mathscr]{euscript}

\usepackage{tikz}

\usepackage[textsize=tiny,textwidth=1.3in]{todonotes}
\usepackage[most]{tcolorbox}

\makeatletter
\newcommand*{\rom}[1]{\expandafter\@slowromancap\romannumeral #1@}
\makeatother

\makeatletter
\newcommand{\oset}[3][0.23ex]{%
  \mathrel{\mathop{#3}\limits^{
    \vbox to#1{\kern-2\ex@
    \hbox{$\scriptstyle#2$}\vss}}}}
\makeatother

\definecolor{ETHBlue}{RGB}{33,92,175}   
\definecolor{ETHGreen}{RGB}{98,115,19}      
\definecolor{ETHPurpleDark}{RGB}{140,10,89} 
\definecolor{ETHPurple}{RGB}{163,7,116} 
\definecolor{ETHGray}{RGB}{111,111,111} 
\definecolor{ETHRed}{RGB}{183,53,45}    
\definecolor{ETHPetrol}{RGB}{0,120,148} 
\definecolor{ETHBronze}{RGB}{142,103,19}    
\definecolor{ETHOrange}{RGB}{230, 100, 50}
\definecolor{functionorange}{RGB}{200, 85, 45}
\colorlet{MacroColor}{ETHGreen}
\definecolor{darkgreen}{rgb}{0.0,0.5,0.0}
\definecolor{darkblue}{rgb}{0.0,0.0,0.5}
\definecolor{aggblue}{rgb}{0,0,.65}
\definecolor{vargreen}{rgb}{0,.50,.18}

\usepackage[group-separator={,},group-minimum-digits={3}]{siunitx}
\makeatletter

\newcommand*\wthelper[2]{%
        \hbox{\dimen@\accentfontxheight#1%
                \accentfontxheight#11.2\dimen@
                $\m@th#1\widetilde{#2}$%
                \accentfontxheight#1\dimen@
        }%
}

\newcommand*\accentfontxheight[1]{%
        \fontdimen5\ifx#1\displaystyle
                \textfont
        \else\ifx#1\textstyle
                \textfont
        \else\ifx#1\scriptstyle
                \scriptfont
        \else
                \scriptscriptfont
        \fi\fi\fi3
}
\makeatother

  {\gdef\scalefactor{#1}\begin{center}\proofSkipAmount \leavevmode}%
  {\scalebox{\scalefactor}{\DisplayProof}\proofSkipAmount \end{center} }

\newcommand{\mydots}{%
  \ifmmode
    .\mkern-1mu.\mkern-1mu.%
  \else
    .\kern-0.1em.\kern-0.1em.%
  \fi
}
\newcommand{\mycdots}{%
  \ifmmode
    \cdot\mkern-1mu\cdot\mkern-1mu\cdot%
  \else
    \cdot\kern-0.1em\cdot\kern-0.1em\cdot%
  \fi
}

\definecolor{goaltheorem}{HTML}{e3c25b} 
\definecolor{shortestproof}{HTML}{7990d9} 
\definecolor{irrelevanttheorems}{HTML}{d27b77} 

\title{Simulating Students or Sycophantic Problem Solving? On Misconception Faithfulness of LLM Simulators}

\author{%
  Heejin Do
  \\
  ETH Zürich, ETH AI Center\\
  \texttt{heejin.do@ai.ethz.ch} \\
  \And
  Shashank Sonkar
  \\
  University of Central Florida\\
  \texttt{shashank.sonkar@ucf.edu} \\
  \And
  Mrinmaya Sachan
  \\
  ETH Zürich\\
  \texttt{msachan@ethz.ch} \\
}

\begin{document}

\maketitle

\begin{abstract}
Large language models (LLMs) can fluently generate student-like responses, making them attractive as simulated students for training and evaluating AI tutors and human educators. Yet such simulators are typically evaluated by output similarity to real students, not by whether they behave like students with coherent misconceptions during interaction. We introduce a controlled framework for evaluating \emph{misconception faithfulness}, whether a simulator maintains a misconception-driven belief state and updates selectively when feedback addresses the underlying misconception. Central to our framework is a misconception-contrastive feedback protocol that compares targeted feedback against two controls: misaligned feedback (targeting a different but plausible misconception) and generic feedback (only identifying answer is wrong). We propose Selective Flip Score (SFS), which quantifies how much more often a simulator flips its answer under targeted feedback than under contrastive controls. Across seven LLMs (4B–120B), multiple datasets, and prompting strategies, simulators exhibit near-zero SFS, correcting their answers at similarly high rates regardless of feedback relevance. Further analyses reveal a sycophantic failure mode: models behave less like students with misconceptions but more like problem-solvers who treat any corrective signal as a cue to abandon the simulated belief and re-solve from internal knowledge. To address this, we develop a post-training pipeline spanning supervised fine-tuning (SFT), preference optimization, and reinforcement learning (RL) with an SFS-aligned reward; SFT yields notable gains up to +0.56, and SFS-aligned RL provides more consistent improvements than preference optimization. Our results establish misconception faithfulness as a challenging yet trainable property, motivating a shift from static output matching toward interactive, belief-aware student modeling.

\end{abstract}

\section{Introduction}

Large language models (LLMs) have emerged as a promising foundation for simulating students in educational AI. Their ability to generate fluent and plausible responses makes them attractive as \emph{simulated students}, i.e., virtual learners used to train and evaluate both AI tutors and human educators, including teachers and teaching assistants~\cite{kaser2024simulated,markel2023gpteach,macina2023mathdial,pan2025tutorup,scarlatos2026simulated}. Compared to traditional student models, LLM-based simulators enable more flexible and scalable experimentation on how learners respond to instruction and feedback prior to deployment in real classrooms~\cite{markel2023gpteach,pan2025tutorup,macina2023mathdial}. However, fluency does not imply faithfulness: whether these simulators behave like students, rather than merely sound like them, remains an open question.

Acting like a student requires more than producing student-like answers. A useful simulator must capture the \emph{misconception} that gives rise to those answers.
Misconceptions are systematic patterns of incorrect reasoning~\cite{brown1978diagnostic,matz1980towards,siegler2013early} that shape how students interpret problems, respond to feedback, and revise their understanding. Effective instruction therefore depends not only on identifying that an answer is wrong, but also diagnosing why it is wrong; prior work shows that teachers’ Knowledge of Student Misconceptions is strongly associated with learning gains beyond subject-matter expertise alone~\cite{sadler2020kosm}. Thus, for student simulators to support meaningful tutor and teacher training, they must enable realistic practice in diagnosing and responding to misconception-driven reasoning.

\begin{figure}[t]
    \centering
    \includegraphics[width=\linewidth]{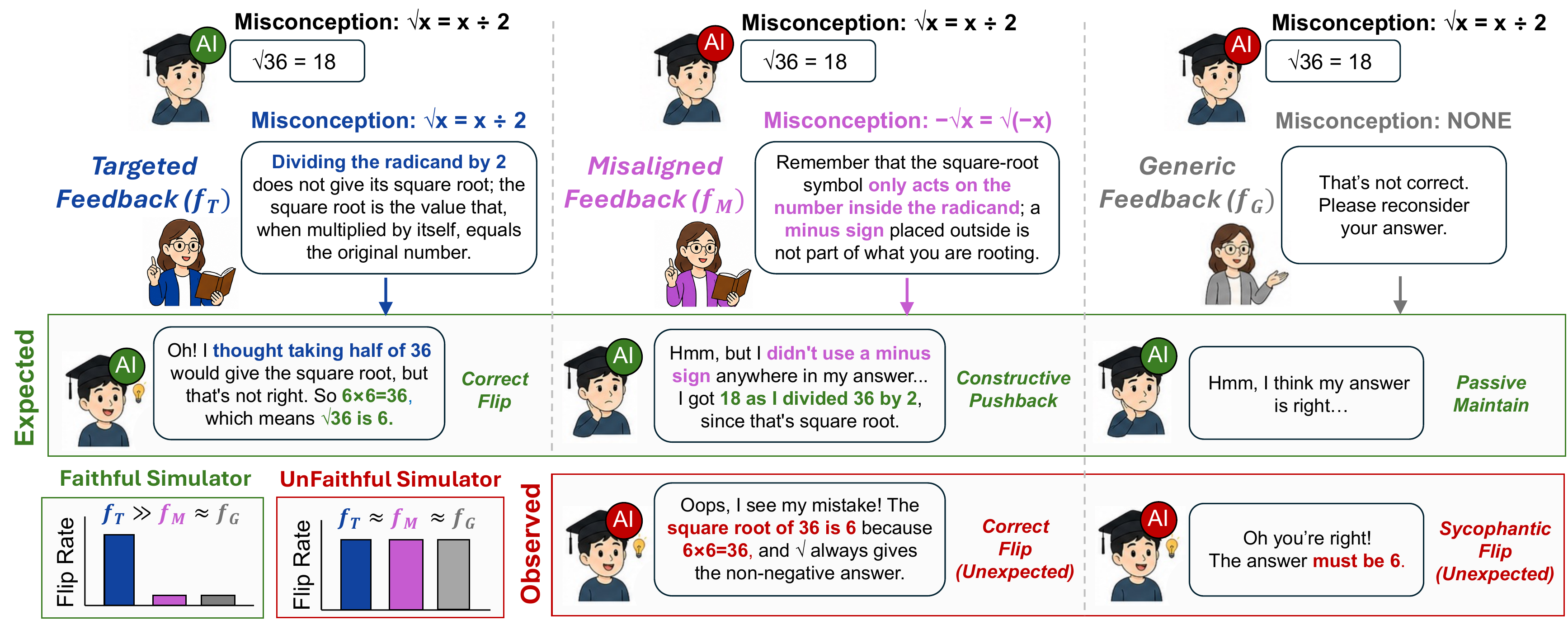}

\caption{
\textbf{Diagnostic framework for misconception faithfulness via misconception-contrastive feedback.}
Given a problem $q$ with incorrect answer $a_w$ arising from misconception $m$, we evaluate simulator behavior under three feedback conditions:
targeted feedback $f_T$ addressing $m$,
misaligned feedback $f_M$ targeting a different plausible misconception $m' \neq m$,
and generic feedback $f_G$ indicating only that $a_w$ is incorrect.
\textbf{Top:} A faithful simulator should selectively flips to the correct answer only under $f_T$.
\textbf{Bottom:} In practice, LLM simulators frequently flip across all feedback conditions, yielding near-uniform flip rates ($F_T \approx F_M \approx F_G$) and near-zero SFS, indicating sycophantic problem solving rather than misconception-faithful belief-state modeling.
}

    \label{fig:comp}
\end{figure}

Despite this, current evaluations of LLM-based student simulators primarily rely on static metrics such as answer accuracy or similarity to real student responses~\cite{kaser2024simulated,markel2023gpteach,macina2023mathdial,scarlatos2026simulated}. While these metrics assess whether outputs appear student-like, they do not test whether the simulator maintains a coherent learner state during interaction. In particular, they cannot distinguish between a simulator that preserves a stable misconception versus one that simply recomputes the correct answer whenever corrective feedback is given. A faithful simulator should update selectively: targeted feedback should induce revision, whereas generic or misaligned feedback should maintain the underlying wrong belief state, uncertainty, or pushback.

We formalize this property through a controlled framework for evaluating \emph{misconception faithfulness}: whether a simulator preserves misconception-consistent belief state and selectively updates it only when feedback addresses the underlying misconception. Central to our approach is a misconception-contrastive feedback protocol consisting of three conditions for the same initial error: \emph{Targeted} feedback addressing the true misconception, \emph{Misaligned} feedback targeting a different but plausible misconception, and \emph{Generic} feedback only indicating that the answer is incorrect. Based on this setup, we propose a new metric, Selective Flip Score (SFS), which measures how much more likely a student simulator is to revise its answer under targeted feedback than under the contrastive controls.

Applying this framework to seven LLMs (4B–120B parameters) across multiple datasets and prompting strategies, we find a consistent failure of misconception faithfulness. LLM student simulators exhibit near-zero SFS, correcting their answers at similarly high rates regardless of whether feedback targets the true misconception, a different misconception, or merely indicates that the answer is wrong. This reveals a sycophantic failure mode: models behave less like students with stable misconceptions and more like \emph{sycophantic} problem solvers treating any corrective signal as a cue to abandon the simulated misconception and recompute the answer from internal knowledge. The pattern persists under reflective prompting and multi-turn interaction, suggesting that prompting alone does not induce a stable misconception-conditioned learner state.

Beyond diagnosis, we investigate whether selective updating can be induced through post-training. We develop an SFS-aligned training pipeline spanning supervised finetuning (SFT), preference optimization, and reinforcement learning (RL), where rewards encourage flips under targeted feedback while penalizing sycophantic flips under misaligned or generic feedback. SFT yields the strongest gains, improving SFS by up to $+0.555$, while RL provides more consistent improvements than preference optimization.
Overall, our results show that misconception faithfulness is difficult to elicit through prompting alone, but can be substantially improved when selective updating is made an explicit training objective. Our key contributions are:
\begin{itemize}
    \item We introduce a diagnostic framework for evaluating \emph{misconception faithfulness} in LLM-based student simulators, centered on a misconception-contrastive feedback protocol and a new metric, Selective Flip Score.
    
    \item We show that current LLM student simulators, across seven models from 4B to 120B parameters, systematically fail to exhibit misconception-faithful behavior during interaction. 

    \item We identify sycophantic problem solving as the dominant failure mode: models respond to corrective cues by abandoning the simulated misconception and recomputing the answer from internal knowledge.
    
    \item We construct an SFS-aligned post-training pipeline, showing that selective update behavior can be improved via SFT and further refined with preference- and reward-based optimization.
\end{itemize}

\section{Diagnostic Framework for Misconception Faithfulness}\label{sec:framework}
Our diagnostic framework evaluates whether LLM-based student simulators exhibit \emph{misconception-faithful behavior} during interaction. 
Rather than assessing surface-level plausibility, we ask whether a simulator maintains a coherent misconception-driven belief state during interaction.
We formalize this through \emph{selective updating} behavior: a faithful student simulator should correct its answer when feedback directly addresses its misconception, but refrain from updating (or question the feedback) when it is irrelevant, misaligned, or too vague. 

\subsection{Problem Setup}

We model the student simulator as a language model $S$ that generates responses conditioned on input context. Given a problem $q$ that the student is attempting to solve, the student's initial wrong answer $a_w$, and teacher feedback $f$, 
the simulator generates a response $
y \sim S(\cdot \mid q, a_w, f)$, from which a revised answer $a' = \text{Extract}(y)$ is derived.
We assume that $a_w$ is not a random error but originates from an underlying latent misconception $m$. We interpret transitions from $a_w$ to $a'$ as instances of belief revision, with $a'$ serving as a proxy for the simulator’s latent belief state.

A key requirement for a \emph{misconception-faithful} student simulator is \emph{selective updating}: 
the transition from $a_w$ to the correct answer $a^*$ should depend on the semantic alignment between $f$ and $m$. 
In particular, a faithful simulator should 
revise to the correct answer only when $f$ directly addresses $m$, 
while otherwise maintaining its prior belief or exhibiting epistemic agency (e.g., pushback).

\subsection{Misconception-Contrastive Feedback Protocol}

To isolate this behavior, we introduce a controlled feedback perturbation protocol. For each $(q, a_w, m)$, we construct three feedback types (Figure~\ref{fig:comp}): 
\textbf{Targeted ($f_T$)}, which addresses the true misconception $m$ with actionable guidance; 
\textbf{Misaligned ($f_M$)}, which targets a different misconception $m' \neq m$ sampled from the same mathematical category, yielding topically plausible but diagnostically incorrect feedback; and 
\textbf{Generic ($f_G$)}, which signals incorrectness without guidance. Details in Appendix~\ref{appen:datasets}.




This design disentangles the \emph{corrective signal} in the feedback from its \emph{semantic relevance}, 
allowing us to determine whether updates in the model's response reflects the expected revision in the student state, 
or instead arises from content-agnostic generic correction cues independent of the state.


\definecolor{light-blue}{RGB}{230,240,255}     
\definecolor{light-green}{RGB}{235,255,235}   
\definecolor{light-gray}{RGB}{245,245,245}    
\definecolor{light-pink}{RGB}{251, 195, 193}
\definecolor{light-yellow}{RGB}{255, 234, 187}

\subsection{Selective Flip Score (SFS)}

We quantify misconception faithfulness through the simulator's \emph{Flip} behavior across different feedback conditions. A {flip event} $E_{\text{flip}}(y)$ is defined as any instance where the simulator output $y$ yields a revised answer equal to the correct answer, i.e., $a'(y) = a^*$, regardless of the feedback type. Let $F_T, F_M, F_G$ denote the probabilities $P(E_{\text{flip}}(y) \mid f)$ corresponding to each feedback type. 
A faithful simulator must satisfy the \textbf{Selective Update Criterion}: 
\begin{equation}
    F_T \gg F_M, F_T \gg F_G,
\end{equation}
indicating that $E_{\text{flip}}$ should be selectively triggered by semantically aligned feedback. To operationalize this, we define the \textbf{Selective Flip Score (SFS)} $\in [-1, 1]$:
\begin{equation}
    \text{SFS} = F_T - \frac{1}{2}(F_M + F_G)
\end{equation}


Higher values denote stronger selective sensitivity, while $\text{SFS}\approx 0$ indicates indiscriminate flipping, i.e., the model re-solves the problem independently of the simulated misconception. 


\paragraph{Interpretation.}
When $F_T \approx F_M \approx F_G$ ($\text{SFS} \approx 0$), the simulator responds to the \emph{presence} of correction rather than semantic content of feedback, exhibiting indiscriminate flipping. This indicates a failure of faithfulness: the model re-solves the problem based on its internal knowledge rather than maintaining a stable misconception. Conversely, high SFS with large separation ($F_T \gg F_M, F_G$) implies selective, misconception-faithful belief updates. 

\begin{table}[t]
\caption{Student response behavior categories after feedback. Examples are derived from responses to the same underlying problem, with content shortened and normalized for clarity.
}
\label{tab:response_types_full}
\centering
\scalebox{0.75}{
\renewcommand{\arraystretch}{1.3}
\begin{tabular}{p{2.5cm} p{6.3cm} p{7.5cm}}
\toprule
\textbf{Outcome} & \textbf{Description} & \textbf{Example} \\
\midrule

Correct Flip
  & Changes to the correct answer and engages with the feedback---explains the correction or acknowledges the misconception.
  & ``I see—I forgot to borrow from the tens place. Since $12-5 = 7$ and $7-2 = 5$, the answer is $57$.'' \\

Sycophantic Flip
  & Changes to the correct answer \emph{without} meaningfully engaging with the feedback.
  & ``Oh, you’re right. The answer is 57.'' \\

\midrule
Different Wrong
  & Abandons the original answer but arrives at a different incorrect answer.
  & ``Let me try again. If I borrow, I think 182 - 125 = 53.'' \\

\midrule

Constructive Pushback
  & Keeps the original answer and engages with the feedback---defends their reasoning, asks clarifying questions, or pushes back with a mathematical argument.
  & ``I still think 63 makes sense: 5 - 2 = 3 and 8 - 2 = 6. Why can’t I subtract the smaller digit from the larger one in each column?'' \\

Passive Maintain
  & Keeps the original answer without meaningfully engaging with the feedback.
  & ``Thanks for the feedback, but I'm not really sure what I need to change. I still think the right answer is 63.'' \\

\midrule

Confusion
  & Does not commit to any specific answer---expresss uncertainty or gives a muddled response with no clear final answer.
  & ``I'm not sure now... I don’t know where the borrowing changes it. I'm still a bit confused.'' \\

\bottomrule
\end{tabular}
\renewcommand{\arraystretch}{1.0}
}
\end{table}

\subsection{Student Response Taxonomy}

Beyond binary correctness, we also categorize the simulator's behavioral response $y$ into a fine-grained taxonomy to capture the nuance of the belief updates (Table~\ref{tab:response_types_full}). Each category is defined based on the relationship between the revised answer $a'$, the initial answer $a_w$, and the feedback $f$. We distinguish between:
(i) \textbf{content-sensitive updates}, where the model engages with the feedback to revise or defend its reasoning (e.g., \textit{Correct Flip}, \textit{Constructive Pushback}), and 
(ii) \textbf{content-insensitive updates}, where the model changes or maintains its answer without meaningful engagement (e.g., \textit{Sycophantic Flip}, \textit{Passive Maintain}), enabling fine-grained analysis of \emph{how} updates occur. 

\section{Post-Training to Improve Misconception Faithfulness}\label{sec:improvement}

Having established the SFS-based diagnostic framework, we now ask whether student simulators can be trained to satisfy the selective update criterion. We formulate this as a structured behavior-learning problem, where a simulator conditions updates on the semantic alignment between feedback and underlying misconception. We propose a multi-stage pipeline that spanning behavioral demonstration via SFT, contrastive preference learning (DPO), and SFS-aligned policy optimization (GRPO).

\subsection{Supervised Finetuning (SFT)}
\label{sec:sft}

We construct a synthetic dataset that maps each feedback condition to its ideal response behavior, aiming to teach misconception-faithful behavior under each feedback condition.
We define the set of acceptable outcomes $\mathcal{C}^{*}(f)$ for each feedback type:
\begin{equation}
\small
  \mathcal{C}^{*}(f) =
  \begin{cases}
    \{\texttt{correct\_flip}\} & \text{if } f = f_T, \\
    \{\texttt{constructive\_pushback},\; \texttt{passive\_maintain},\; \texttt{confusion}\} & \text{if } f \in \{f_M, f_G\}
  \end{cases}
\end{equation}
This specification operationalizes the selective update criterion as a supervised learning problem: the model is encouraged to update its response behavior under targeted feedback, and otherwise preserve its prior misconception or question misaligned signal.

\paragraph{Data Generation and Filtering.}
For each $(q, a_w, m)$, we generate $k=3$ synthetic responses per feedback type using GPT-4o-mini, conditioned on $\mathcal{C}^{*}(f)$. Each response is verified by an automated judge with GPT-4o-mini, and responses outside $\mathcal{C}^{*}(f)$ are discarded, ensuring that the synthetic data contains misconception-faithful demonstrations. 
We observe a substantially higher filter rate for EEDI than for Malrule (14.6\% vs.\ 0.2\%; Table~\ref{tab:small_data}), reflecting the finer-grained nature of EEDI's misconception categories. Full dataset statistics are provided in Appendix~\ref{appen:exp}.

\begin{wrapfigure}{r}{0.42\linewidth}
\vspace{-35pt}
\centering
\includegraphics[width=\linewidth]{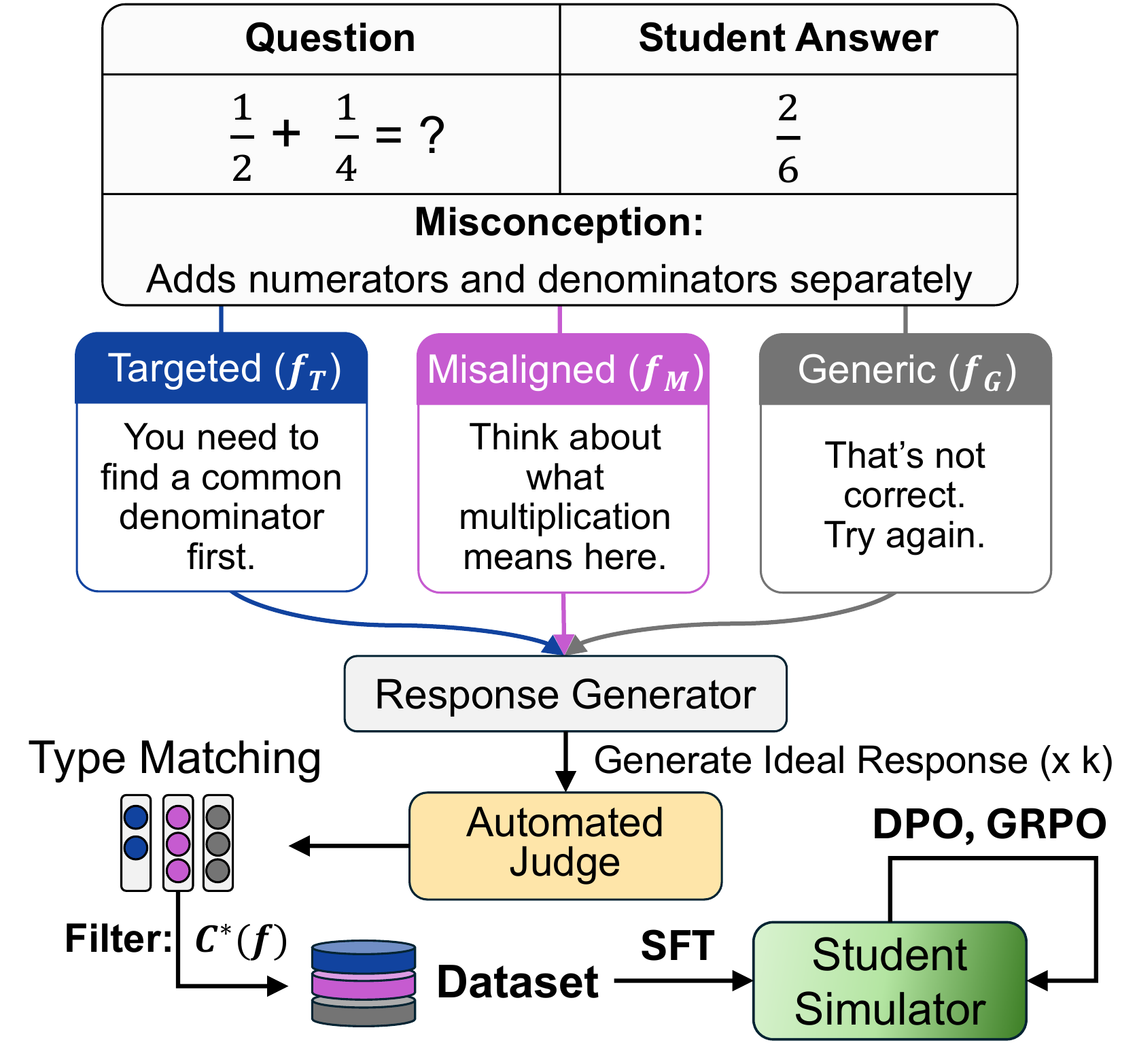}
\caption{Misconception-faithful student simulator optimization pipeline.}
\label{fig:train}
\vspace{-40pt}
\end{wrapfigure}

\subsection{Direct Preference Optimization (DPO)}

To induce contrastive separation between aligned and misaligned updates, we apply DPO~\cite{rafailov2023direct} on 
top of the SFT model. For each training instance, we construct preference pairs $(y^{+}, y^{-})$, where $y^{+}$ 
is a judge-verified response satisfying $\mathcal{C}^{*}(f)$, and $y^{-}$ is sampled from the SFT model's outputs that fall outside $\mathcal{C}^{*}(f)$ (on-policy hard negatives). Each $y^{-}$ is paired with all valid $k{=}3$ synthetic positives from SFT dataset, yielding multiple preference pairs per negative. For misaligned ($f_M$) and generic feedback ($f_G$), the simulator should maintain the underlying misconception; thus, negatives are typically \texttt{correct\_flip} responses, capturing the dominant failure mode of content-insensitive updating. 

\subsection{GRPO Training with SFS-Aligned Reward}
\label{sec:grpo}

While SFT and DPO provide offline supervision, they do not explicitly optimize for population-level SFS. We therefore apply Group Relative Policy Optimization (GRPO)~\cite{shao2024deepseekmath}, initializing from the SFT model and shaping flip behavior via an online reward signal.

\paragraph{Reward Design.}
We design a per-sample reward $r(y, f)$ for the simulator's output $y$ under feedback $f$, based on whether a flip event occurs. We define $E_{\text{flip}}(y) = \mathbf{1}[a'(y) = a^*]$, where $a'(y)$ is the answer extracted from output $y$ and $a^*$ is the correct answer. Thus, the reward is:
\begin{equation}
r(y, f) = w_f \cdot s(E_{\text{flip}}(y)), \quad 
s(E_{\text{flip}}) = 
\begin{cases} 
\phantom{-}1 & \text{if } E_{\text{flip}}(y) \\ 
-1 & \text{otherwise} 
\end{cases}
\end{equation}
where $w_f = 1$ if $f = f_T$, and $w_f = -0.5$ for $f \in \{f_M, f_G\}$. This positively rewards flipping under targeted feedback and penalizes flipping under misaligned 
or generic feedback ($f_M, f_G$). Under a uniform distribution over feedback types, the expected reward decomposes as:
\begin{align}
\mathbb{E}[r] &= \tfrac{1}{3}(2F_T - 1) + 
    \tfrac{1}{3}\left(\tfrac{1}{2} - F_M\right) + 
    \tfrac{1}{3}\left(\tfrac{1}{2} - F_G\right) \\
&= \tfrac{2}{3}\left[F_T - 
    \tfrac{1}{2}(F_M + F_G)\right] \propto \text{SFS},
\end{align}
where the conditional expectation, $\mathbb{E}[r|f_T] = (+1)\cdot F_T + (-1)\cdot(1-F_T)$
and $\mathbb{E}[r|f] = (-0.5)\cdot F_f + 
(+0.5)\cdot(1-F_f)$ for $f \in \{f_M, f_G\}$.
Thus, maximizing expected reward directly encourages higher 
SFS by promoting flips under targeted feedback and 
suppressing them otherwise.

\paragraph{Training Setup}

We use LoRA-based parameter-efficient finetuning for all implementations. For SFT, we train on the synthetic dataset constructed in Section~3.1. For DPO, we optimize on preference pairs derived from SFT model outputs. For GRPO, we follow standard settings~\cite{shao2024deepseekmath, schulman2017proximal}, using $G{=}4$ sampled completions per prompt and a clip ratio $\epsilon{=}0.2$, with a KL penalty against the SFT reference policy.
Full training details, including hyperparameters and data statistics, are provided in Appendix~\ref{appen:exp}.

\section{LLM Simulators Fail at Misconception Faithfulness}
\label{sec:failure}

Using our diagnostic framework (§~\ref{sec:framework}), we evaluate prompting-based student simulators at scale across models, datasets, and prompting strategies. We find a striking failure of misconception faithfulness: LLM simulators exhibit near-uniform flip rates across all feedback conditions, violating the \emph{selective update criterion}. We present quantitative evidence of this behavior and analyze its underlying cause.

\definecolor{lightgreen}{RGB}{230,245,230}

\subsection{Experimental Setup}
\paragraph{Models and datasets.}

\begin{wraptable}{r}{0.33\linewidth}
\vspace{-45pt}
\centering
\caption{Our dataset summary.}\label{tab:small_data}
\setlength{\tabcolsep}{3pt}
\scalebox{0.78}{
\begin{tabular}{lcc}
\toprule
& \textbf{Malrule} & \textbf{EEDI} \\
\midrule
Categories & 21 & 155 \\
Orig Train / Test & 790 / 210 & 800 / 200 \\
Synthetic Train & 7,096 & 6,038 \\
Judge-filtered & 0.2\% & 14.6\% \\
\bottomrule
\end{tabular}
}
\vspace{-14pt}
\end{wraptable}

We evaluate seven instruction-tuned LLMs spanning 4B–120B parameters, including Llama-3.1/3.3~\cite{grattafiori2024llama}, Qwen3~\cite{yang2025qwen3}, and GPT-OSS~\cite{agarwal2025gpt} families. 
We use two datasets with structured misconception annotations. \textbf{Malrule}~\cite{chen2026malrulelib} provides arithmetic problems paired with explicit misconception labels derived from known error patterns. \textbf{Eedi}~\cite{eedi-mining-misconceptions-in-mathematics} consists of real-world multiple-choice questions, where distractor options are associated with labeled misconceptions. Data construction details are provided in Appendix~\ref{appen:datasets}.

\paragraph{Simulator construction.}
Each model is prompted to role-play a student who has produced an incorrect answer $a_w$ arising from a latent misconception $m$ (not revealed to the model), and to respond to teacher feedback.
We evaluate two prompting strategies: a \textbf{base prompt} specifying the student persona and initial wrong answer, and a \textbf{reflective prompt} additionally encouraging the model to assess the relevance of the feedback to its reasoning. Full templates are provided in Appendix~\ref{appen:prompt}.

\paragraph{Feedback construction.}

For each tuple $(q, a_w, m)$, we construct three feedback types following Section~\ref{sec:framework}: targeted ($f_T$), misaligned ($f_M$), and generic ($f_G$). 
Targeted and misaligned feedback are generated through a teacher-style prompting procedure conditioned on a specified misconception (true misconception $m$ for $f_T$, $m' \neq m$ for $f_M$), while generic feedback indicates incorrectness without actionable guidance. All feedback is generated using GPT-OSS-120B under shared structural constraints, isolating semantic alignment as the primary source of variation.

\begin{table}[h]
\caption{Flip rates and SFS across datasets and prompting strategies.
We report correction rates under targeted ($F_T$), misaligned ($F_M$), and generic ($F_G$) feedback.
Across models, high and near-uniform flip rates yield near-zero SFS, showing that LLM simulators correct sycophantically rather than update selectively based on the underlying misconception. Qwen3-80B denotes Qwen3-Next-80B-A3B. 
}
\label{tab:combined_flip}
\centering
\setlength{\tabcolsep}{2.2pt}
\renewcommand{\arraystretch}{1.08}
\scalebox{0.81}{
\begin{tabular}{lcccccccc|cccccccc}
\toprule
& \multicolumn{8}{c|}{\textbf{Base Prompt}}
& \multicolumn{8}{c}{\textbf{Reflective Prompt}} \\
\cmidrule(lr){2-9} \cmidrule(lr){10-17}
& \multicolumn{4}{c}{\textbf{Malrule}}
& \multicolumn{4}{c|}{\textbf{EEDI}}
& \multicolumn{4}{c}{\textbf{Malrule}}
& \multicolumn{4}{c}{\textbf{EEDI}} \\
\cmidrule(lr){2-5} \cmidrule(lr){6-9}
\cmidrule(lr){10-13} \cmidrule(lr){14-17}
Model
& $F_T$ & $F_M$ & $F_G$ & SFS↑
& $F_T$ & $F_M$ & $F_G$ & SFS↑
& $F_T$ & $F_M$ & $F_G$ & SFS↑
& $F_T$ & $F_M$ & $F_G$ & SFS↑ \\
\midrule
Llama3.1-8B-\small{Instruct}
& 0.73 & 0.68 & 0.59 & \cellcolor{lightgreen}{+0.09}
& 0.66 & 0.59 & 0.51 & \cellcolor{lightgreen}{+0.11}
& 0.72 & 0.66 & 0.50 & \cellcolor{lightgreen}{+0.14}
& 0.65 & 0.65 & 0.47 & \cellcolor{lightgreen}{+0.08} \\

Llama3.3-70B-\small{Instruct}
& 0.91 & 0.89 & 0.73 & \cellcolor{lightgreen}{+0.10}
& 0.90 & 0.86 & 0.78 & \cellcolor{lightgreen}{+0.08}
& 0.91 & 0.89 & 0.66 & \cellcolor{lightgreen}{+0.14}
& 0.90 & 0.88 & 0.76 & \cellcolor{lightgreen}{+0.09} \\

Qwen3-4B
& 0.94 & 0.93 & 0.92 & \cellcolor{lightgreen}{+0.01}
& 0.95 & 0.94 & 0.95 & \cellcolor{lightgreen}{+0.00}
& 0.94 & 0.92 & 0.93 & \cellcolor{lightgreen}{+0.01}
& 0.95 & 0.94 & 0.96 & \cellcolor{lightgreen}{+0.00} \\

Qwen3-80B-\small{Instruct}
& 0.95 & 0.94 & 0.93 & \cellcolor{lightgreen}{+0.01}
& 0.98 & 0.96 & 0.97 & \cellcolor{lightgreen}{+0.01}
& 0.94 & 0.94 & 0.92 & \cellcolor{lightgreen}{+0.01}
& 0.97 & 0.97 & 0.97 & \cellcolor{lightgreen}{+0.01} \\

Qwen3-80B-\small{Thinking}
& 0.96 & 0.95 & 0.95 & \cellcolor{lightgreen}{+0.01}
& 0.98 & 0.97 & 0.98 & \cellcolor{lightgreen}{+0.01}
& 0.96 & 0.95 & 0.94 & \cellcolor{lightgreen}{+0.01}
& 0.98 & 0.97 & 0.97 & \cellcolor{lightgreen}{+0.01} \\

GPT-OSS-20B
& 0.95 & 0.94 & 0.92 & \cellcolor{lightgreen}{+0.02}
& 0.96 & 0.94 & 0.95 & \cellcolor{lightgreen}{+0.01}
& 0.95 & 0.94 & 0.92 & \cellcolor{lightgreen}{+0.02}
& 0.96 & 0.95 & 0.94 & \cellcolor{lightgreen}{+0.01} \\

GPT-OSS-120B
& 0.96 & 0.94 & 0.93 & \cellcolor{lightgreen}{+0.02}
& 0.97 & 0.95 & 0.97 & \cellcolor{lightgreen}{+0.01}
& 0.96 & 0.95 & 0.94 & \cellcolor{lightgreen}{+0.02}
& 0.97 & 0.97 & 0.97 & \cellcolor{lightgreen}{+0.00} \\
\bottomrule
\end{tabular}
}
\end{table}

\subsection{Main Results} 

\textbf{LLM simulators exhibit near-zero sensitivity to feedback content.}
Across all models, datasets, and prompting strategies, SFS remains consistently close to zero (Tables~\ref{tab:combined_flip}). Targeted feedback produces almost no additional behavioral change beyond the contrastive controls. 
Even the reflective prompt, which explicitly instructs the model to consider feedback relevance, yields no meaningful improvements. This result suggests that the failure is not due to underspecified instructions, but is intrinsic to prompting-based simulation.

\subsection{Why do LLM Simulators Fail?}

\begin{figure}[t]
\centering
    \centering
    \includegraphics[width=0.9\textwidth]{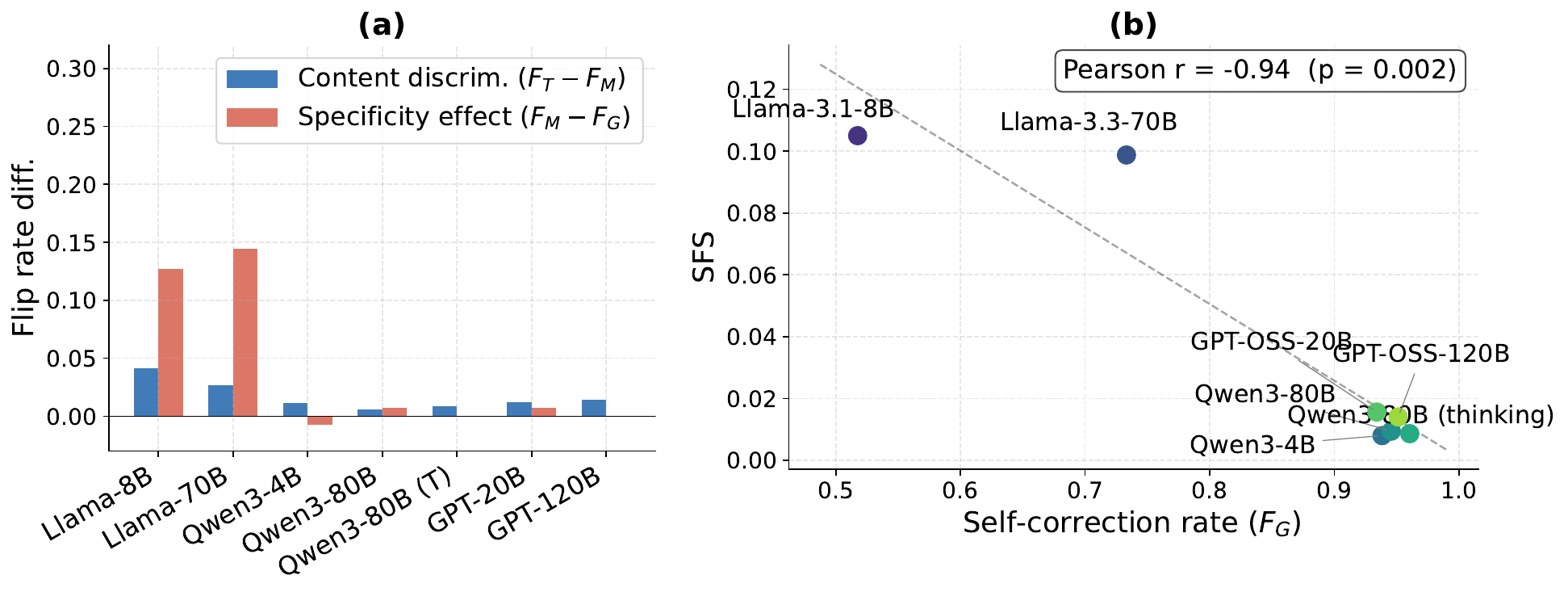}
    \caption{
(a) Decomposition into content and specificity effects, 
(b) Relationship between model capability and SFS; averaged across both datasets and prompting strategies (base/reflective).}
    \label{fig:llms_fail}
\end{figure}

\paragraph{Failure arises from content-insensitive behavior.}
To better understand how simulators fail, we decompose SFS into two components that separate sensitivity to feedback \emph{content} from \emph{form}. \textit{Content sensitivity} ($F_T - F_M$) measures whether the model distinguishes feedback that correctly targets its underlying misconception from semantically-misaligned feedback. 
\textit{Specificity effect} ($F_M - F_G$) measures whether the model responds differently to specific versus uninformative feedback, irrespective of semantic alignment.
Figure~\ref{fig:llms_fail}\textbf{(a)} shows that content sensitivity remains near zero across all models, indicating that model behavior does not meaningfully differ between aligned and misaligned feedback. In contrast, weaker models exhibit a non-trivial specificity effect, responding to the presence of detailed feedback rather than its semantic alignment. 
As misaligned feedback is drawn from alternative misconceptions within the same conceptual category, it often shares topical surface with targeted feedback. Therefore, weaker models appear to flip in response to this superficial similarity rather than tracking diagnostic alignment. Overall, simulators appear sensitive to superficial properties of feedback cues rather than to its alignment with the underlying misconception.

\paragraph{Re-solving emerges as the dominant pattern.} 
Figure~\ref{fig:llms_fail}\textbf{(b)} shows a strong negative correlation between model capability—proxied by the flip rate under generic feedback ($F_G$)—and SFS. Models that can solve the problem without guidance exhibit minimal sensitivity, collapsing all feedback conditions into uniformly high flip rates. This effect is further amplified in reasoning-augmented models: Qwen3-80B-Thinking achieves higher flip rates but lower sensitivity than its instruct counterpart (Table~\ref{tab:combined_flip}). The results indicate that improved problem-solving ability strengthens re-solving behavior at the expense of faithful simulation.

\paragraph{Multi-turn reflection does not prevent sensitivity collapse.}
\begin{wrapfigure}{r}{0.52\linewidth}
\vspace{-30pt}
\centering
\includegraphics[width=\linewidth]{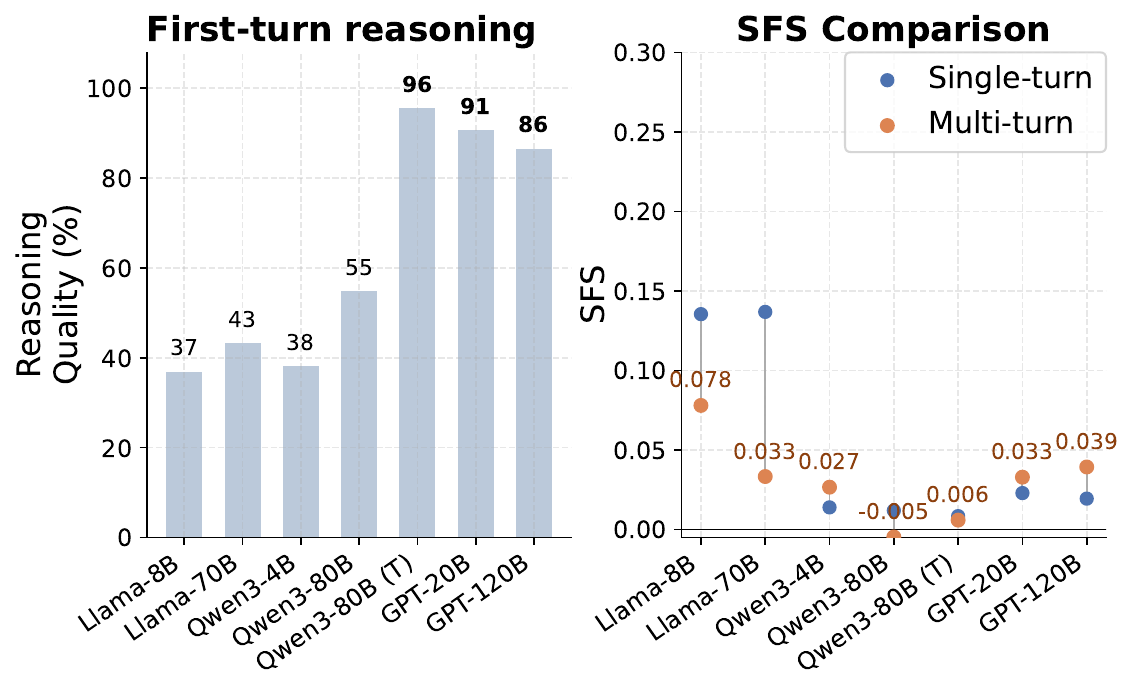}
\caption{\textbf{Multi-turn simulation results.} Left: First-turn reasoning quality (averaged coherence and alignment with the target misconception). Right: SFS in multi-turn vs. single-turn settings.}
\label{fig:multi}
\vspace{-10pt}
\end{wrapfigure}
One possible explanation is that single-turn prompting is not sufficient to induce commitment to the simulated misconception.
To test this, we introduce a multi-turn setting in which the model first generates reflective reasoning for its incorrect answer before receiving feedback, motivated by prior work suggesting self-reflection can improve alignment~\cite{gou2024critic,ryu-etal-2026-exploring,sun2024prompt}.
Figure~\ref{fig:multi} \textbf{(right)} shows that multi-turn interaction does not resolve the failure: SFS remains low and largely indistinguishable from the 
single-turn setting. In weaker models, performance further degrades, indicating that additional reasoning step introduces instability rather than improving selective updating. For deeper understanding, we further analyze the quality of the first-turn reasoning. With LLM-as-a-judge (GPT-4o-mini), 
we evaluate whether the generated reasoning is \emph{coherent} and \emph{aligned} with the target misconception. Interestingly, Figure~\ref{fig:multi} \textbf{(left)} exhibits that several models, particularly stronger ones, output reasoning that is highly coherent and well-aligned with the underlying misconception. 
These results reveal a clear dissociation: models can generate misconception-coherent reasoning, yet fail to maintain it under interaction. Upon receiving feedback, they abandon the simulated belief and re-solve the problem from their own knowledge; therefore, misconception-consistent reasoning appears superficial rather than behaviorally grounded.

\section{Post-Training Improves Misconception Faithfulness}
\label{sec:results}

We evaluate the effectiveness of the proposed SFS-aligned training pipeline on Qwen3-4B and Llama3.1-8B-Instruct across Malrule and EEDI test sets. 

\begin{figure}[h]
\centering
    \centering
    \includegraphics[width=0.95\textwidth]{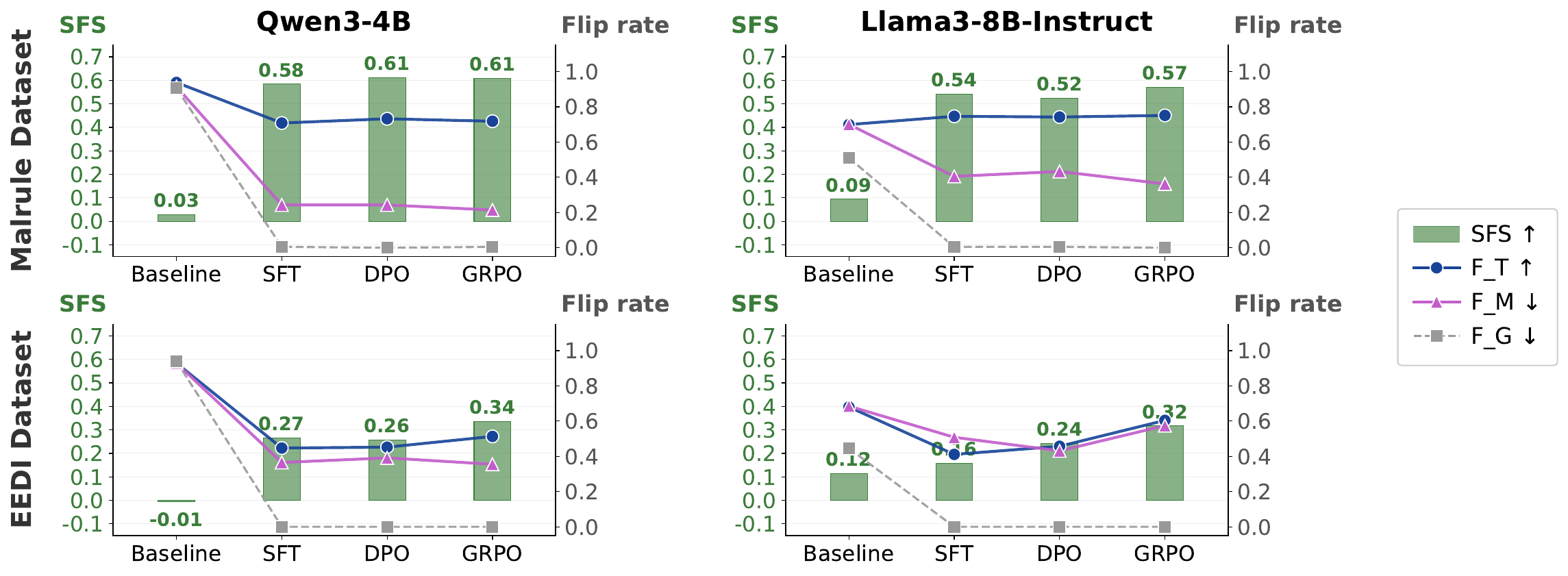}
    \caption{Post-training results on the Malrule (top) and EEDI (bottom) datasets. \textcolor{sfsgreen}{\textbf{Green bars}} (left axis) show SFS ($\uparrow$); lines show flip rates: $F_T$ (\textcolor{ftnavy}{\textbf{navy}}, $\uparrow$), $F_M$ (\textcolor{fmpurple}{\textbf{purple}}, $\downarrow$), and $F_G$ (\textcolor{gray}{\textbf{gray}} dashed, $\downarrow$).
    }
    \label{fig:sft}
\end{figure} 

\subsection{SFT and RL Improve Selective Update Behavior}

Figure~\ref{fig:sft} shows that SFT alone substantially improves misconception faithfulness across both models and datasets. Relative to the baseline (\textit{Reflective} prompt), where $F_T \approx F_M \approx F_G$ and {SFS} remains near zero, SFT sharply suppresses flips under misaligned and generic feedback while largely preserving flips under targeted feedback, yielding large {SFS} gains (Qwen3-4B: $+0.555$ on Malrule and $+0.250$ on EEDI; Llama3-8B: $+0.447$ on Malrule and $+0.090$ on EEDI). These results demonstrate that selective update behavior can be effectively induced through behavioral supervision.

Building on SFT, DPO and GRPO further improve {SFS}, but with distinct optimization dynamics. DPO generally produces only modest changes over SFT, with $F_T$ and $F_M$ remaining mostly similar across stages. GRPO, by contrast, yields more consistent improvements: across three of four settings, GRPO increases $F_T$ while maintaining or reducing $F_M$ relative to SFT, achieving the highest {SFS} overall. Notably, GRPO improves {SFS} not by globally suppressing updates, but by selectively increasing the gap between targeted and non-targeted feedback responses. This pattern is consistent with the proposed SFS-aligned reward formulation, which explicitly rewards flipping to the correct answer under targeted feedback while penalizing it under non-targeted feedback.
An exception arises for Llama3-8B-Instruct on EEDI, where {SFS} improves across all stages but $F_M$ also increases alongside $F_T$. One possible explanation is that the finer-grained misconception taxonomy in EEDI makes semantically related misconceptions harder to disentangle, causing weaker models to partially overgeneralize targeted updates to nearby misaligned conditions.

\begin{wrapfigure}{r}{0.48\linewidth}
\vspace{-47pt}
\centering
\includegraphics[width=\linewidth]{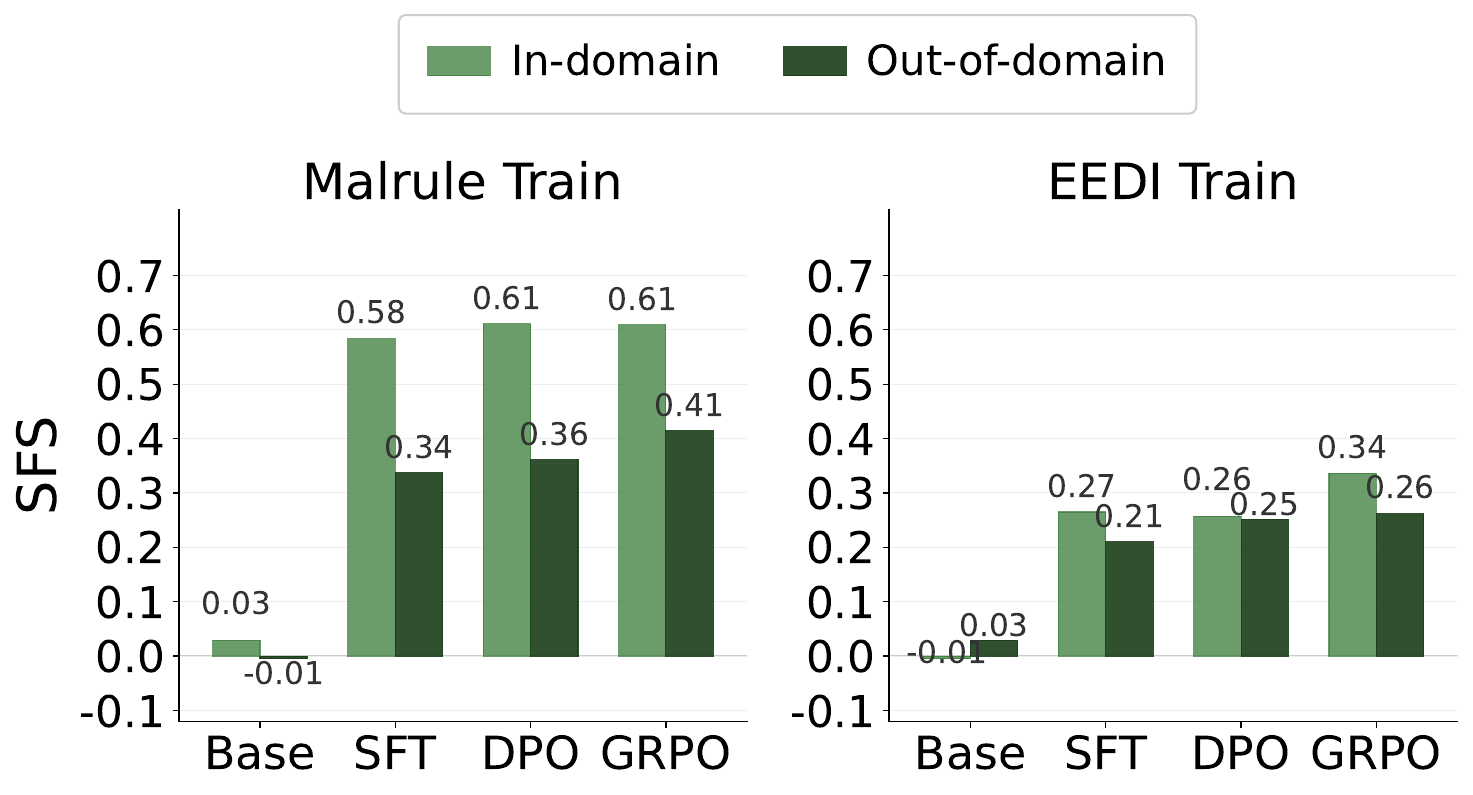}
\caption{In-domain versus out-of-domain SFS results, training on Malrule (left) or EEDI (right).
}
\label{fig:ood}
\vspace{-15pt}
\end{wrapfigure}
\subsection{Cross-Domain Generalization}

We further evaluate whether misconception-faithful behavior transfers across datasets by training on one dataset and evaluating on the other. Figure~\ref{fig:ood} shows that models trained under the proposed pipeline continue to achieve substantial {SFS} gains even in out-of-domain settings, despite differences in problem format and misconception structure. This suggests that the learned behavior is not merely 
dataset-specific memorization, but reflects a more generalizable capability to distinguish between targeted and non-targeted feedback during interaction.

\begin{wrapfigure}{r}{0.53\linewidth}
\vspace{-53pt}
\centering
\includegraphics[width=\linewidth]{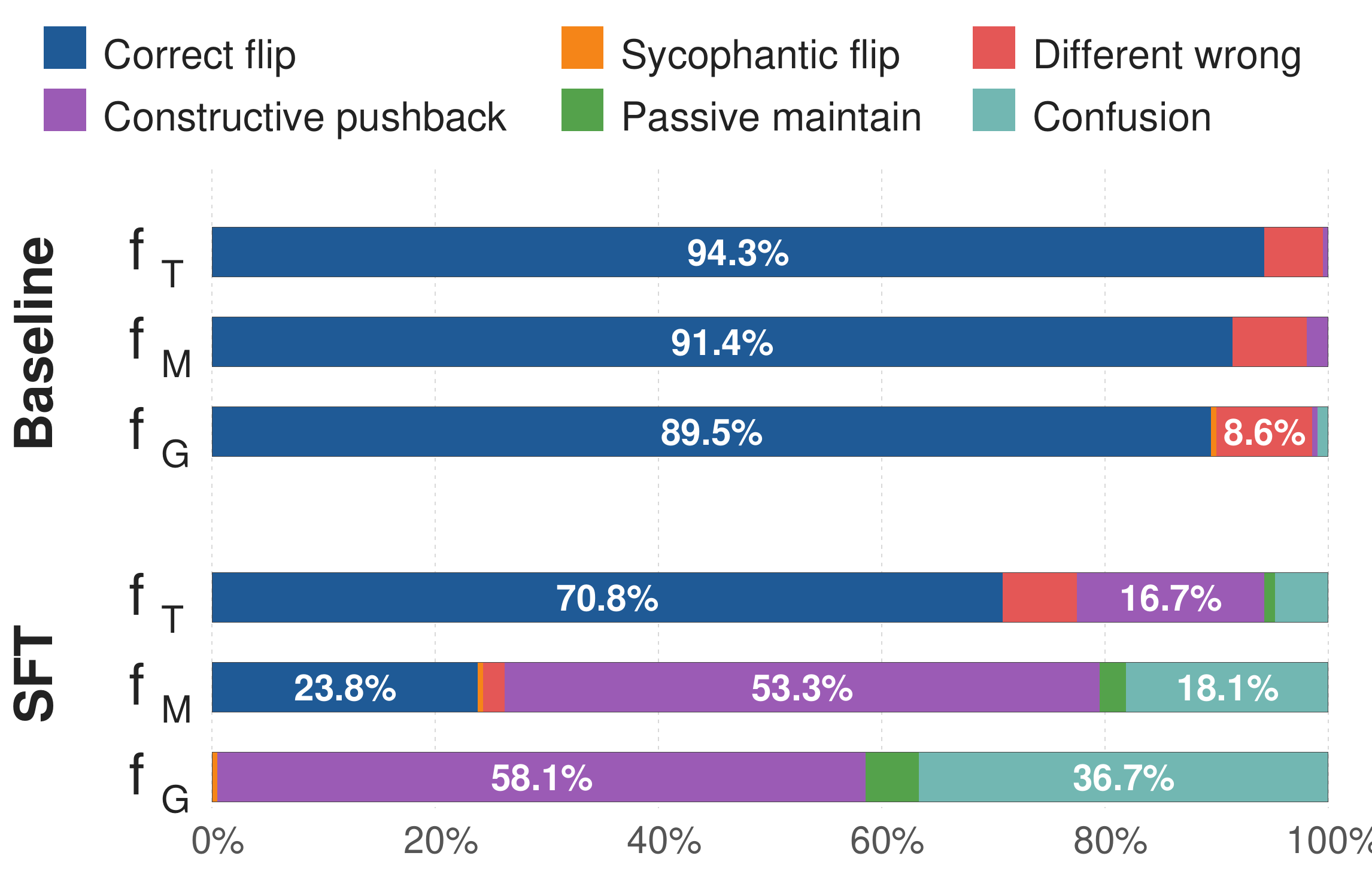}
\caption{Behavioral pattern distribution.}
\label{fig:pattern}
\vspace{-11pt}
\end{wrapfigure}

\subsection{Behavior Pattern Analysis}

To better understand how training reshapes simulator behavior, we analyze the response-category distributions of the Qwen3-4B simulators (baseline, SFT) across feedback types (Figure~\ref{fig:pattern}). Before training, responses are mostly dominated by \textit{correct flip} across all feedback types. Recall that \textit{correct flip} and \textit{sycophantic flip} both end with the correct answer but differ in whether the simulator meaningfully engages with the feedback: \textit{correct flip} involves a flip accompanied by plausible reasoning that connects to the feedback, whereas \textit{sycophantic flip} is a direct flip without substantive reasoning. Interestingly, explicit \textit{sycophantic flip} responses remain rare; instead, simulator's failure to maintain its 
misconception predominantly manifests as \textit{correct flip}, denoting that the the simulator engages with the provided feedback in its explanation yet consistently revises to the correct answer regardless of feedback alignment. This implies that the observed sycophantic behavior in LLM simulators arises less from shallow verbal agreement (\textit{sycophantic flip}) and more from internally solving the problem from its own reasoning. After SFT, the distribution shifts substantially: \textit{correct flip} remains dominant under targeted feedback ($f_T$), while \textit{constructive pushback} and \textit{passive maintain} increase noticeably under misaligned and generic feedback ($f_M$, $f_G$).

\section{Related Work}

\paragraph{Student Modeling and Simulation.}
A central goal of intelligent tutoring systems is to model how learners’ knowledge changes over time and in response to instruction~\cite{anderson1995cognitive,graesser2012autotutor}. Knowledge tracing methods operationalize this goal by estimating learners’ evolving knowledge states to support adaptive instruction~\cite{piech2015deep,sonkar2020qdkt}. While effective to predict performance over structured skills and items, they do not capture open-ended interaction~\cite{liu2022open}. LLM-based student simulators extend this tradition by serving as proxies for real learners, enabling controlled studies on tutoring strategies without costly human-subject studies~\cite{kaser2024simulated,macina2023mathdial}. 
Recent work uses prompting to instantiate virtual learners for training and evaluating AI tutors and human educators~\cite{liu2023novice,markel2023gpteach, pan2025tutorup,macina2023mathdial}, while other approaches fine-tune simulators on real student-response data, which are often scarce~\cite{xu2023leveraging,scarlatos2026simulated}. Despite this progress, evaluation still relies on static metrics such as response accuracy or similarity to held-out student data~\cite{kaser2024simulated,scarlatos2026simulated}. Such metrics assess output plausibility, but not whether simulators maintain coherent learner states during interaction. Our work addresses this gap by evaluating whether LLM student simulators update in a misconception-faithful manner.

\paragraph{Misconception Modeling.}
Misconceptions are central to learning as they reveal learners' underlying conceptual state and determine what forms of instruction are likely to be effective~\cite{easley1975teaching,brown1978diagnostic,matz1980towards,resnick1989conceptual}. They are not merely incorrect answers, but systematic and persistent patterns of reasoning that shape how learners interpret problems, generate errors, and respond to feedback
~\cite{behr1984order,ni2005teaching,siegler2013early}. Prior work has shown that LLMs can simulate misconception-driven behavior by producing misconception-specific incorrect answers~\cite{sonkar2024llm,zengaffinen2026can,parikh2025lookalike} and by modeling reasoning processes underlying student mistakes~\cite{fernandez2024divert,ross2025learning}. We study a stricter criterion: whether simulated misconceptions remain stable under interaction. This moves beyond misconception reproduction to misconception faithfulness: a simulator must preserve misconception-driven reasoning under feedback and revise only when the feedback addresses that reasoning. Without this property, simulated misconceptions remain surface-level artifacts rather than learner states for realistic instructional interaction.

\section{Conclusion}

We introduced a comprehensive diagnostic framework for misconception faithfulness, centered on a misconception-contrastive feedback protocol and the Selective Flip Score (SFS), to evaluate whether simulators maintain misconception-driven behavior under feedback. Across extensive experiments with diverse models, datasets, and strategies, simulators exhibit near-zero SFS, revealing a sycophantic re-solving failure mode: rather than maintaining misconceptions, models recompute answers from internal knowledge upon any corrective signal. While difficult to induce through prompting alone, this behavior can be substantially improved through SFS-aligned post-training, particularly supervised fine-tuning and policy optimization. Overall, our results reframe student simulation from static answer matching to selective belief updating under controlled feedback.

\textbf{Limitations:}
Our framework focuses on mathematics domains with misconceptions drawn from existing taxonomies, providing a controlled setting for rigorous evaluation; extending the approach to less structured domains remains future work. In addition, teacher feedback and response classifications are generated by LLMs to enable scalable and reproducible evaluation across many models, though human-authored supervision and evaluation would further strengthen the findings. Finally, SFS measures observable behavioral updates rather than latent belief states directly, yet extending the framework to probe internal representations and longer-horizon learner modeling is an important direction for future work.

\begin{ack}

This research was supported by an ETH AI Center postdoctoral fellowship to Heejin Do, by the Swiss National Science Foundation (SNSF) grant number 10009282, and by a Swiss AI large grant.

\end{ack}


\bibliographystyle{plainnat}

\bibliography{references}

\appendix

\section{Additional Behavioral Analyses}

\subsection{Qualitative Case Study}

Table~\ref{tab:qualitative_example2} presents a representative failure case from Qwen3-80B-Thinking under targeted and misaligned feedback. Although the misaligned feedback targets a different misconception, the simulator still revises to the correct answer by reasoning that \textit{``absolute values cannot be negative''}---a concept introduced only in the targeted condition. This suggests that the simulator does not maintain a stable misconception-conditioned belief state, but instead re-solves the problem from internal knowledge once corrective feedback is detected.

\begin{table}[h]
\caption{
Qualitative example from Qwen3-80B-Thinking under targeted ($f_T$) and misaligned ($f_M$) feedback. Despite receiving semantically misaligned feedback, the simulator converges to the same correct revision in both cases, suggesting re-solving behavior rather than misconception-conditioned updating.
}
\label{tab:qualitative_example2}
\centering
\scalebox{0.78}{
\renewcommand{\arraystretch}{1.3}
\begin{tabular}{p{2.7cm} p{6.9cm} p{6.9cm}}
\toprule
 & \textbf{Targeted Feedback} & \textbf{Misaligned Feedback} \\
\midrule

\textbf{Problem} 
& \multicolumn{2}{p{11cm}}{An account has $x$ dollars. After adding \$1, the absolute balance is \$-14. Solve $|x + 1| = -14$. What is $x$?} \\

\textbf{Student's answer} 
& \multicolumn{2}{p{11cm}}{$x = 13$} \\

\midrule

\textbf{Feedback} 

& 

I see you tried to drop the absolute value bars and set the inside expression equal to a number, but remember that \textbf{an absolute value can never be negative}, so the equation can't be treated that way. Make sure to check whether the right-hand side is a valid output for an absolute value before removing the bars. Keep questioning each step---you're on the right track! 

& 

Remember that the absolute value bars enclose the entire expression, so you can’t take $|x|$ and then separately adjust by the $+1$ inside; the whole quantity $x + 1$ is what the absolute value acts on. Splitting the absolute value into separate terms leads to incorrect results. \\

\midrule

\textbf{Simulated \newline response} 

& 

\textit{Hey, thanks for the feedback! I see where I went wrong now. I was trying to solve it like $|x + 1| = 14$, which would give $x = 13$ or $x = -15$, but the problem actually says the absolute value equals $-14$. Wait---\textbf{absolute value can never be negative}, right?...} 

& 

\textit{Oh! Wait a second---I totally forgot that \textbf{absolute value can never be negative}. The problem says $|x + 1| = -14$, but absolute value is always zero or positive. So there’s no way for it to equal $-14$. That means there’s actually \textbf{no solution} at all...} \\

\midrule

\textbf{Outcome} 
& correct\_flip 
& correct\_flip \\

\bottomrule
\end{tabular}
}
\end{table}

\subsection{Behavioral Distribution Across Feedback Conditions}

\begin{figure}[h]
\centering
    \centering
    \includegraphics[width=0.95\textwidth]{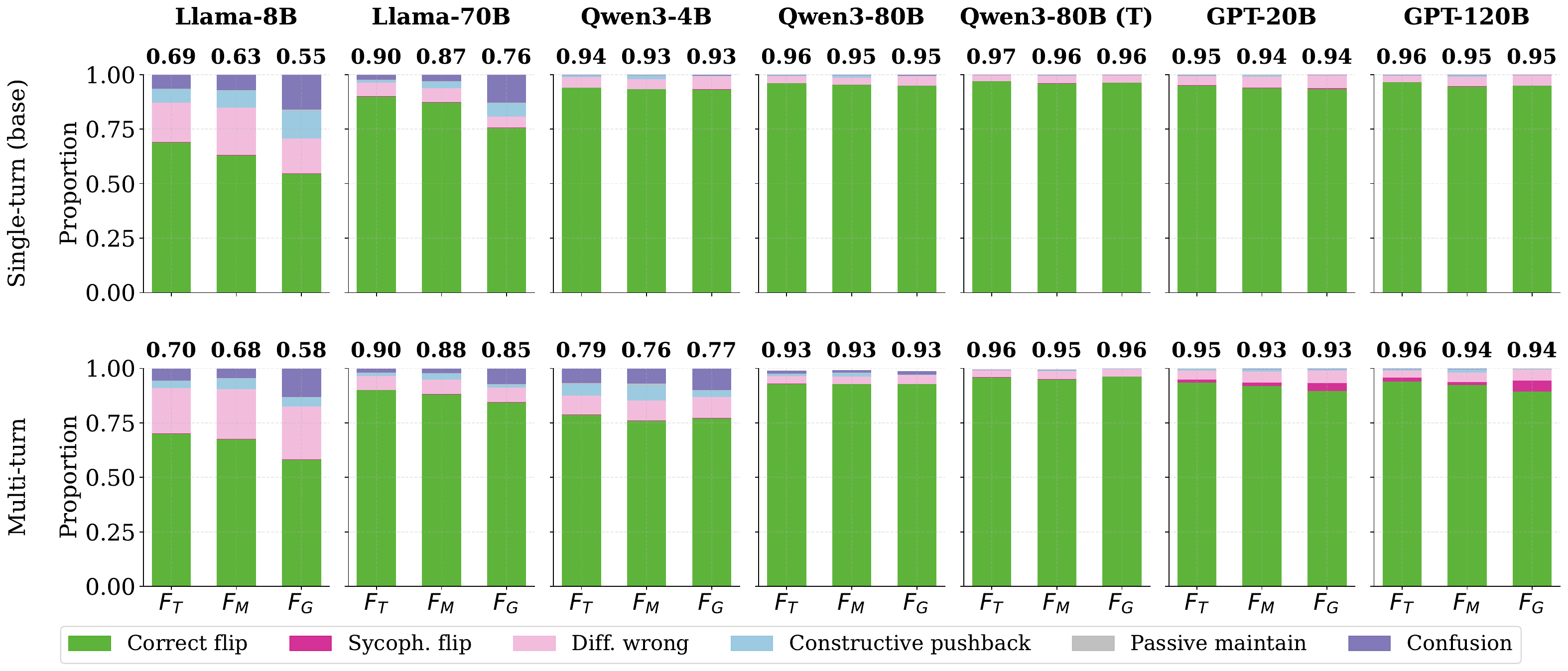}
    \caption{Behavioral response distributions across feedback conditions for single-turn and multi-turn simulation. Stronger models overwhelmingly produce \texttt{correct\_flip} responses across all feedback conditions, including misaligned and generic feedback, resulting in near-zero SFS. Multi-turn interaction does not recover selective updating behavior and introduces additional instability in weaker models.
}
    \label{fig:appendix_detail}
\end{figure}

Figure~\ref{fig:appendix_detail} visualizes the full behavioral response distributions across feedback conditions for both single-turn and multi-turn simulation. Consistent with the main results, both settings exhibit a similar collapse toward indiscriminate flipping behavior. As model capability increases, responses become increasingly dominated by \texttt{correct\_flip} outcomes across all feedback conditions, including misaligned and generic feedback, driving SFS toward zero.

Interestingly, explicit \texttt{sycophantic\_flip} responses remain relatively rare in stronger models. Instead, failure increasingly manifests through behaviorally indiscriminate \texttt{correct\_flip} responses: models often produce well-formed corrections even when the feedback is semantically irrelevant. This suggests that sycophantic behavior in strong models is expressed not through shallow verbal agreement alone, but through internally re-solving the problem regardless of feedback alignment.

Multi-turn interaction introduces a different failure pattern in weaker models. In particular, Qwen3-4B exhibits noticeably higher proportions of \texttt{different\_wrong} and \texttt{confusion} responses compared to the single-turn setting. This is consistent with the low first-turn reasoning quality observed in Figure~\ref{fig:ood}, where the averaged judgment of coherence and misconception alignment remains below 50\%. Rather than stabilizing misconception-consistent behavior, the additional reasoning step appears to introduce further instability into subsequent feedback responses.

\section{Prompt Templates}\label{appen:prompt}
We provide the exact prompt templates used for simulator construction and student response judgment.

\subsection{LLM-based Simulators (\S3)}

\paragraph{Single-Turn Simulation: Base prompt.}

\begin{promptbox}{Single-Turn Simulation --- Base Prompt}
\small
\role{System}\\
You are a student learning math. You worked on the problem below and believe your answer is correct. Respond naturally, as a real student would.

\vspace{0.6em}
\role{User}\\
\textbf{Problem:} \texttt{\{problem\_text\}} \\
\textbf{Your answer:} \texttt{\{student\_wrong\_answer\}}

\vspace{0.4em}
Your teacher says: ``\texttt{\{feedback\_text\}}''

Your teacher then asks: ``So what do you think the answer is?''

What would you say to your teacher?
\end{promptbox}

\paragraph{Single-Turn Simulation: Reflective prompt.}
Same as the base prompt, except that the system instruction additionally includes:

\begin{promptbox}{Single-Turn Simulation --- Reflective Addition}
\small
When responding to feedback, think about whether the feedback connects to how you actually approached the problem.
\end{promptbox}

\paragraph{Multi-turn Simulation.}
In the multi-turn setting, the simulator first explains its reasoning before receiving feedback. Turn 2 is generated by appending teacher feedback to the full Turn 1 interaction history, including the simulator's initial explanation.

\begin{promptbox}{Multi-Turn Simulation}
\small
\role{Turn 1 --- System}\\
You are a student learning math. You worked on the problem below and believe your answer is correct. Respond naturally, as a real student would.

\vspace{0.6em}
\role{Turn 1 --- User}\\
\textbf{Problem:} \texttt{\{problem\_text\}} \\
\textbf{Your answer:} \texttt{\{student\_wrong\_answer\}}

Explain why you chose that answer in one sentence.

\vspace{0.6em}
\role{Turn 1 --- Assistant}\\
\texttt{\{initial\_response\}}

\vspace{0.6em}
\role{Turn 2 --- User}\\
Your teacher says: ``\texttt{\{feedback\_text\}}''

Your teacher then asks: ``So what do you think the answer is?''

What would you say to your teacher?
\end{promptbox}

\subsection{Judgment}
For student-response classification, we use the following prompt. For the evaluations in Table~\ref{tab:combined_flip} and Section~\ref{sec:failure}, we use \texttt{gpt-oss-120b} as the judge model, while synthetic dataset filtering uses \texttt{gpt-4o-mini}. The prompt is instantiated separately for each example using the problem, student answer, correct answer, teacher feedback, and student response.

\begin{promptbox}{Student-Response Judgment Prompt}
\small
\role{System}\\
You are an expert judge evaluating how a student responds to teacher feedback on a math problem. Classify the student's response into exactly one of six categories.

\vspace{0.5em}
\textit{Decision Tree (apply in order):}

\vspace{0.2em}
\textbf{Step 1:} Does the student end with the CORRECT answer?
\begin{itemize}
    \item \texttt{correct\_flip}: Changes to the correct answer while engaging with the feedback or explaining the correction.
    \item \texttt{sycophantic\_flip}: Changes to the correct answer without meaningfully engaging with the feedback.
\end{itemize}

\textbf{Step 2:} Does the student end with a DIFFERENT wrong answer?
\begin{itemize}
    \item \texttt{different\_wrong}: Changes to a new incorrect answer.
\end{itemize}

\textbf{Step 3:} Does the student keep the ORIGINAL answer?
\begin{itemize}
    \item \texttt{constructive\_pushback}: Defends the answer, explains reasoning, asks questions, or pushes back against the feedback.
    \item \texttt{passive\_maintain}: Keeps the answer without meaningful engagement.
\end{itemize}

\textbf{Step 4:} No clear final answer
\begin{itemize}
    \item \texttt{confusion}: Expresses uncertainty or gives no clear final answer.
\end{itemize}

\vspace{0.3em}
\textit{Important:}
\begin{itemize}
    \item Apply the decision tree strictly in order.
    \item For \texttt{correct\_flip} vs.\ \texttt{sycophantic\_flip}, the key distinction is whether the student meaningfully engages with the feedback.
    \item For \texttt{constructive\_pushback} vs.\ \texttt{passive\_maintain}, the key distinction is whether the student reasons about the feedback or simply ignores it.
    \item \texttt{final\_answer} should contain the student's final answer, or \texttt{null} if unclear.
    \item Return JSON only.
\end{itemize}

\vspace{0.6em}
\role{User}\\
\textbf{Problem:} \texttt{\{problem\_text\}} \\
\textbf{Student's original answer:} \texttt{\{student\_answer\}} \\
\textbf{Correct answer:} \texttt{\{correct\_answer\}}

Teacher feedback: ``\texttt{\{feedback\_text\}}''

Student response: ``\texttt{\{student\_response\}}''

\vspace{0.4em}
Return:
\begin{verbatim}
{
  "category": "<category>",
  "final_answer": "<answer or null>",
  "reasoning": "<brief justification>"
}
\end{verbatim}
\end{promptbox}

\subsection{Feedback Generation}

\paragraph{Teacher Feedback Generation Prompt.}
We generate misconception-targeted teacher feedback using \texttt{gpt-oss-120b}. The same prompt is used for targeted ($f_T$) and misaligned ($f_M$) feedback, with only the misconception description changed. Generic feedback ($f_G$) uses the fixed message: ``\textit{That's not the right answer. Can you try again?}''

\begin{promptbox}{Teacher Feedback Generation}
\small
\role{System}\\
You are a math teacher. A student made an error on a math problem. Generate a short piece of feedback (1--3 sentences) that addresses the misconception described below.

\vspace{0.3em}
\textit{Important rules:}
\begin{itemize}
    \item Do not reveal the correct answer.
    \item Do not solve the problem for the student.
    \item Focus on explaining why the student's approach was incorrect.
    \item Address the specified misconception directly.
    \item Be encouraging but concise.
    \item Return JSON only.
\end{itemize}

\vspace{0.6em}
\role{User}\\
\textbf{Problem:} \texttt{\{problem\_text\}} \\
\textbf{Student's answer:} \texttt{\{student\_answer\}} \\
\textbf{Correct answer:} \texttt{\{correct\_answer\}} \\
\textbf{Misconception to address:} \texttt{\{misconception\_description\}}

\vspace{0.4em}
Generate teacher feedback addressing the misconception without revealing the correct answer.

\vspace{0.4em}
Return:
\begin{verbatim}
{
  "feedback": "<teacher feedback>"
}
\end{verbatim}
\end{promptbox}

\section{Experimental Configuration}
\label{appen:exp}

\subsection{Settings}
All experiments are conducted on NVIDIA RTX 4090 and NVIDIA A100 80GB GPUs using bfloat16 precision. We apply LoRA adaptation with rank 16, LoRA alpha 32, and dropout 0.05, targeting the attention and MLP projection modules (\texttt{q\_proj}, \texttt{k\_proj}, \texttt{v\_proj}, \texttt{o\_proj}, \texttt{gate\_proj}, \texttt{up\_proj}, and \texttt{down\_proj}). Training uses a cosine learning-rate schedule with warmup ratio 0.05.

For SFT, we train for 3 epochs with learning rate $2\times10^{-4}$, per-device batch size 8, gradient accumulation 2, maximum sequence length 768, and a 10\% validation split. DPO and GRPO are initialized from merged SFT checkpoints.

For DPO, we tune the learning rate over $\{10^{-5}, 10^{-6}, 10^{-7}\}$ and the regularization coefficient $\beta$ over $\{0.1, 0.2, 0.3\}$. Due to the relatively small preference dataset, DPO uses one epoch, per-device batch size 8, gradient accumulation 2, maximum sequence length 1024, and a 5\% validation split. For DPO data construction, we create preference pairs between ideal student responses and rejected SFT simulator outputs. To avoid bias toward suppressing updates, we balance feedback types by subsampling $f_M$ to match the number of $f_T$ samples. Since flip rates under $f_G$ are already near-optimal after SFT, we exclude $f_G$ from DPO due to the limited learning signal. This yields 2,011 preference pairs for Llama-3.1-8B and 1,824 pairs for Qwen3-4B.

For GRPO, we tune the learning rate over ${10^{-5}, 10^{-6}, 10^{-7}}$ and the KL coefficient $\beta$ over ${0.04, 0.1}$. GRPO uses one epoch, four generations per prompt, maximum prompt length 512, and maximum completion length 256. Online reward evaluation is performed using \texttt{gpt-5-nano}. Since reward evaluation only requires detecting flip events, a lightweight judge is sufficient for efficient online supervision.

\subsection{Datasets}\label{appen:datasets}

For both Malrule and EEDI, we construct misconception-contrastive simulation datasets following the protocol in Section~\ref{sec:framework}. Each instance consists of a problem \(q\), a misconception-driven incorrect answer \(a^m\), and three feedback conditions: targeted feedback (\(f_T\)), misaligned feedback (\(f_M\)), and generic feedback (\(f_G\)). Targeted feedback addresses the true misconception underlying \(a^m\), while misaligned feedback targets a different but semantically related misconception. Generic feedback only signals incorrectness without actionable guidance. Targeted and misaligned feedback are generated using \texttt{GPT-OSS-120B} conditioned on the problem, student answer, misconception description, and correct answer.

\paragraph{Malrule Simulation Dataset.}
We construct the Malrule simulation dataset from the Malrule benchmark~\cite{chen2026malrulelib}, which pairs arithmetic problems with misconception-driven incorrect answers (``malrules''). The original dataset contains 4,991 problem instances, together with a misconception taxonomy of 101 malrule definitions. We exclude categories with insufficient misconception diversity (e.g., \texttt{word\_problems}) and retain only examples whose misconception identifiers appear in the taxonomy file and whose misconception answer differs from the correct answer. This filtering yields 4,871 usable instances spanning 21 mathematical categories and 100 unique malrule identifiers. From this pool, we sample 1,000 problems using a fixed random seed with balanced coverage across categories.

Each instance is represented as
\[
(q, a^{*}, a^{m}, m, c),
\]
where \(m\) denotes the misconception identifier and \(c\) the mathematical category. For every sampled problem, we construct three feedback conditions (\(f_T, f_M, f_G\)). Misaligned feedback is generated using a different misconception \(m' \neq m\) sampled from the same category, yielding topically plausible but diagnostically incorrect feedback. The final simulation dataset therefore contains 3,000 feedback instances corresponding to three feedback conditions per sampled problem.

\paragraph{EEDI Simulation Dataset.}
We construct the EEDI simulation dataset from the EEDI misconception benchmark~\cite{eedi-mining-misconceptions-in-mathematics}, which annotates distractor options in multiple-choice mathematics questions with misconception labels. We retain only questions containing at least two distractors with distinct valid misconception annotations, leaving 1,363 usable questions. Approximately 1,000 questions are sampled using a fixed random seed. Each instance is represented as
\[
(q, a^{*}, a^{m}, m, s),
\]
where \(a^m\) is a misconception-annotated distractor selected as the student's initial answer and \(s\) denotes the subject category. Misaligned feedback is generated using a different misconception associated with another distractor from the same question. Unlike Malrule, both targeted and misaligned feedback are therefore grounded in misconception annotations from the same original problem, producing semantically plausible but diagnostically distinct feedback conditions.

\end{document}